\documentclass[10pt]{scrartcl}

\usepackage{amssymb}
\usepackage{amsmath}
\usepackage{amscd}
\usepackage{amsthm}
\usepackage{graphicx}
\usepackage{caption}
\usepackage{subcaption}
\usepackage{url}
\usepackage{algorithm}
\usepackage[noend]{algpseudocode}

\newtheorem{definition}{Definition}[section]
\newtheorem{theorem}[definition]{Theorem}

\newtheorem{proposition}[definition]{Proposition}
\newtheorem{lemma}[definition]{Lemma}

\renewcommand{\qed}{\hspace*{\fill} $\blacksquare$}

\newcommand{\commentout}[1]{}

\newcommand{\introskip}{\vspace{3ex}}

\newcommand{\R}{\mathbb{R}}                    

\newcommand{\args}[1]{\mathop{\left( #1 \right)}} 
\newcommand{\inner}[1]{\mathop{\left\langle #1 \right\rangle}}
\newcommand{\norm}[1]{\mathop{\left\lVert #1 \right\rVert}}
\newcommand{\cbrace}[1]{\mathop{\left\{ #1 \right\}}}
\newcommand{\bracket}[1]{\mathop{\left[ #1 \right]}}

\newcommand{\absS}[2]{\mathop{\left\lvert #1 \right\rvert#2}} 
\newcommand{\argsS}[2]{\mathop{\left( #1 \right)#2}} 

\newcommand{\normS}[2]{\mathop{\left\lVert #1 \right\rVert#2}}

\DeclareMathOperator{\id}{id}                  		

\renewcommand{\S}[1]{{\mathcal{#1}}}           	
\def\vec#1{\mathchoice{\mbox{\boldmath$\displaystyle#1$}}
{\mbox{\boldmath$\textstyle#1$}}
{\mbox{\boldmath$\scriptstyle#1$}}
{\mbox{\boldmath$\scriptscriptstyle#1$}}}

\renewenvironment{cases}{%
\left\{\begin{array}{c@{\quad : \quad}l}}%
{%
\end{array}\right.}

\newcounter{part_counter}
\renewenvironment{part}{
\refstepcounter{part_counter}

\medskip

\noindent
\textbf{\arabic{part_counter}.}%
}%

\begin{document}

\title{The Mean Partition Theorem of Consensus Clustering}
\author{Brijnesh J.~Jain \\
 Technische Universit\"at Berlin, Germany\\
 e-mail: brijnesh.jain@gmail.com}
 
\date{}
\maketitle

\begin{abstract} 
To devise efficient solutions for approximating a mean partition in consensus clustering, Dimitriadou et al.~\cite{Dimitriadou2002} presented a necessary condition of optimality for a consensus function based on least square distances. We show that their result is pivotal for deriving interesting properties of consensus clustering beyond optimization. For this, we present the necessary condition of optimality in a slightly stronger form in terms of the Mean Partition Theorem and extend it to the Expected Partition Theorem.  To underpin its versatility, we show three examples that apply the Mean Partition Theorem: (i) equivalence of the mean partition and optimal multiple alignment, (ii) construction of profiles and motifs, and (iii) relationship between consensus clustering and cluster stability.  
\end{abstract} 

\section{Introduction}
Clustering is a standard technique for exploratory data analysis that finds applications across different disciplines such as computer science, biology, marketing, and social science. The goal of clustering is to group a set of unlabeled data points into several clusters based on some notion of dissimilarity. Inspired by the success of classifier ensembles, consensus clustering has emerged as a research topic \cite{Ghaemi2009,VegaPons2011}. Consensus clustering first generates several partitions of the same dataset. Then it combines the sample partitions to a single consensus partition. The assumption is that a consensus partition better fits to the hidden structure in the data than individual partitions. 

One standard approach of consensus clustering combines the sample partitions to a mean partition \cite{Dimitriadou2002,Domeniconi2009,Filkov2004,Franek2014,Gionis2007,Li2007,Strehl2002,Topchy2005,VegaPons2010}. A mean partition best summarizes the sample partitions with respect to some (dis)similarity function.

In \cite{Dimitriadou2002} Dimitriadou et al.~presented a necessary condition of optimality for a consensus function based on least square distance method on hard (crisp) as well as soft (fuzzy) partitions. Their result has been solely applied for devising efficient algorithms for approximating a mean partition \cite{Dimitriadou2002,Hornik2008}. 

In this contribution, we show that the necessary condition of optimality proposed in \cite{Dimitriadou2002} is pivotal for deriving other interesting results in consensus clustering. For this, we restate and extend the necessary condition of optimality to obtain the Mean and Expected Partition Theorem.  Then we present three results that apply the Mean Partition Theorem:
\begin{enumerate}
\item \emph{Optimal Multiple Alignment}: We show that the problem of computing a mean partition and the problem of finding an optimal multiple alignment are equivalent. This equivalence is inspired by results from the equivalence between multiple sequence alignment and consensus sequences from computational biology \cite{Gusfield1997}. Equivalence of the mean partition to multiple alignment provides access to techniques and algorithms from computational biology and sets the stage for gaining further insight into consensus clustering.
\item \emph{Profiles and Motifs}: As an example of techniques from computational biology, we introduce profiles and motifs as tools to analyze and visualize the results of a cluster ensemble. Profiles provide a statistic about the occurrence of a data point in a cluster for a given multiple alignment. Motifs are subsets of data points for which there is a high consensus that they belong to the same cluster. 
\item \emph{Cluster Stability}: We show that consensus clustering and cluster stability are related via multiple alignments, which in turn is equivalent to the mean partition problem in consensus clustering. 
\end{enumerate}
The Expected Mean Partition Theorem together with the consistency result presented in \cite{Jain2016} forms the basis to extend the finite-sample results to asymptotic results.
The proposed results indicate that the Mean Partition Theorem provides access to ideas, concepts, and techniques from computational biology and can  be useful for analyzing quality properties of consensus clustering.

\section{Background}

Throughout this contribution, we assume that $\S{Z}= \cbrace{z_1, \ldots, z_m}$ is a set of $m$ data points and $\S{C} = \cbrace{c_1, \ldots, c_\ell}$ is a set of $\ell$ cluster labels.

\subsubsection*{Partitions and their Representations}

Partitions usually occur in two forms, in a labeled and in an unlabeled form, where labeled partitions can be regarded as representations of unlabeled partitions. 

\medskip

We begin with describing labeled partitions. By $[0,1]^{\ell \times m}$ we denote the set of all ($\ell \times m$)-matrices with elements from the interval $[0,1]$. Consider the set 
\[
\S{X} = \cbrace{\vec{X} \in [0,1]^{\ell \times m} \,:\, \vec{X}^T\vec{1}_\ell = \vec{1}_m},
\]
where $\vec{1}_\ell \in \R^\ell$ and $\vec{1}_m \in \R^m$ are vectors of all ones. The set $\S{X}$ consists of all non-negative matrices whose rows sum to one. Any matrix $\vec{X} \in\S{X}$ is a labeled partition of $\S{Z}$, because the ordering of the rows imposes a labelling of the clusters. The elements $x_{kj}$ of $\vec{X} = \args{x_{ij}}$ describe the degree of membership of data point $z_j$ to the cluster with label $c_k$. Then the columns $\vec{x}_{:j}$ of $\vec{X}$ represent the data points $z_j$ and the rows $\vec{x}_{k:}$ of $\vec{X}$ represent the clusters $c_k$.

\medskip

Next, we describe unlabeled partitions. For this, observe that the rows of a labeled partition $\vec{X}$ describe a cluster structure. Permuting the rows of $\vec{X}$ results in a labeled partition $\vec{X}'$ with the same cluster structure but with a possibly different labelling of the clusters. In clustering, the particular labelling of the clusters is usually meaningless. What matters is the abstract cluster structure represented by a labeled partition. Since there is no natural labelling of the clusters, we define the corresponding unlabeled partition that abstracts from the labelling. An unlabeled partition is the equivalence class of labeled partitions obtained from one another by relabelling the clusters. Formally, an unlabeled partition $X$ corresponding to a labeled partition $\vec{X} \in \S{X}$ is defined by $X = \cbrace{\vec{PX} \,:\, \vec{P} \in \Pi^\ell}$, where $\Pi^\ell$ is the set of all ($\ell \times \ell$)-permutation matrices. 

The definition of an unlabeled partition as an equivalence class of labeled partitions shows that every labeled partition is a representative of a labeled one. To keep the terminology simple, we briefly call $X$ a \emph{partition}, if $X$ is an unlabeled partition. Moreover, any labeled partition $\vec{X}' \in X$ is called a \emph{representation} of $X$, henceforth. By $\S{P}$ we denote the set of all (unlabeled) partitions with $\ell$ clusters over $m$ data points. Since some clusters may be empty, the set $\S{P}$ also contains partitions with less than $\ell$ clusters. Thus, we consider $\ell \leq m$ as the maximum number of clusters we encounter. Finally, the map
\[
\pi: \S{X} \rightarrow \S{P}, \quad \vec{X} \mapsto \pi(\vec{X}) = X
\]
is the natural projection that sends labeled partitions to their corresponding unlabeled partitions. In other words, $\pi$ sends matrices to partitions they represent. 

\medskip

Though we are only interested in unlabeled partitions, we still need labeled partitions for two reasons: (1) Computers can not easily and efficiently cope with unlabeled partitions unless the clusters carry labels in terms of number or names. (2) Using labeled partitions considerably simplifies derivation of theoretical results.

\subsubsection*{Intrinsic Metric}

We endow the set $\S{P}$ of partitions with an intrinsic metric $\delta$ related to the Euclidean distance such that $(\S{P}, \delta)$ becomes a geodesic space. The Euclidean norm for matrices $\vec{X} \in \S{X}$ is defined by
\[
\norm{\vec{X}}= \argsS{\sum_{k = 1}^\ell \sum_{j = 1}^m \absS{x_{kj}}{^2}}{^{1/2}}.
\]
The Euclidean norm induces a distance function on $\S{P}$ defined by
\[
\delta: \S{P} \times \S{P} \rightarrow \R, \quad (X, Y) \mapsto \min \cbrace{\norm{\vec{X} - \vec{Y}}\,:\, \vec{X} \in X, \vec{Y} \in Y}.
\]
Then the pair $\args{\S{P}, \delta}$ is a geodesic metric space \cite{Jain2015c}, Theorem 2.1. 

\subsubsection*{Representations in Optimal Position}  
Suppose that $X$ and $X'$ are two partitions. Then 
\begin{align}\label{eq:delta<=norm}
\delta(X, X') \leq \norm{\vec{X}-\vec{X}'}
\end{align}
for all representations $\vec{X} \in X$ and $\vec{X}' \in X'$. For some pairs of representations $\vec{X}_{\!*} \in X$ and $\vec{X}'_{\!*} \in X'$ equality holds in Eq.~\eqref{eq:delta<=norm}. In this case, we say that representations $\vec{X}$ and $\vec{X}'$ are in optimal position. Note that pairs of representations in optimal position are not uniquely determined.

\section{Representation Theorems for Partitions}

This section presents the Mean Partition Theorem and the Expected Partition Theorem. For this, we introduce the consensus function as a special case of a Fr\'echet function \cite{Frechet1948}. Using the terminology of Fr\'echet functions links consensus clustering to the field of Non-Euclidean statistics \cite{Bhattacharya2012}. 

\subsection{The Mean Partition Theorem}

The Fr\'{e}chet function of a sample $\S{S}_n = \args{X_1, \ldots, X_n} \in \S{P}^n$ of $n$ partitions is a function of the form 
\begin{align*}
F_n: \S{P} \rightarrow \R, \quad Z \mapsto \frac{1}{n}\sum_{i=1}^n \delta\!\argsS{X_i, Z}{^2}.
\end{align*}
The minimum of $F_n$ exists but is not unique, in general \cite{Jain2015c}. A mean partition of $\S{S}_n$ is any partition $M\in \S{P}$ satisfying
\[
M = \arg\min_{X \in \S{P}} F_n(X).
\]
The Mean Partition Theorem states that any representation $\vec{M}$ of a local minimum $M$ of $F_n$ is the standard mean of sample representations in optimal position with $\vec{M}$.

\begin{theorem}[Mean Partition Theorem]\label{theorem:MPT}
Let $\S{S}_n = \args{X_1, \ldots, X_n} \in \S{P}^n$ be a sample of $n$ partitions. Suppose that $M \in \S{P}$ is a local minimum of the Fr\'echet function $F_n(Z)$ of $\S{S}_n$.  Then every representation $\vec{M}$ of $M$ is of the form
\begin{align}\label{eq:theorem:MPT:eq01}
\vec{M} = \frac{1}{n} \sum_{i=1}^n \vec{X}_{\!i},
\end{align}
where the $\vec{X}_{\!i} \in X_i$ are representations in optimal position with $\vec{M}$.
\end{theorem}

\medskip

The Mean Partition Theorem is a special case of the same theorem for the mean of a sample of attributed graphs \cite{Jain2015a}. Any partition can be regarded as an attributed graph without edges. Nodes represent clusters and node attributes describe the membership values of the data points. 

Dimitiradou et al.~in \cite{Dimitriadou2002} showed that Eq.~\eqref{eq:theorem:MPT:eq01} is a necessary condition of optimality. They did not explicitly stress the (perhaps obvious) property that the representations $\vec{X}_{\!i}$ of the sample partitions $X_i$ are in optimal position with $\vec{M}$. This property is however important for gaining further theoretical insight. 

\subsection{The Expected Partition Theorem}

This section presents the Expected Partition Theorem, which is the analogue of the Mean Partition Theorem for expected Fr\'{e}chet functions. Both theorems are statistically related by 
consistency results presented in \cite{Jain2016}.

\introskip

We assume that $Q$ is a probability measure on $\S{P}$. The function 
\[
F_Q: \S{P} \rightarrow \R, \quad Z \mapsto \int_{\S{P}} \delta(X, Z)^2\, dQ(X)
\]
is the expected Fr\'{e}chet function of $Q$. As for the sample Fr\'echet function $F_n$, the minimum of the expected Fr\'{e}chet function $F_Q$ exists but but is not unique, in general \cite{Jain2015c}. Any partition $M\in \S{P}$ that minimizes $F_Q$ is an expected partition of $Q$. 

\medskip

To state the Expected Partition Theorem in a similar flavor as the Mean Partition Theorem, we need to introduce some concepts. For details, we refer to Section \ref{app:sec:partition-spaces}. Let $\vec{Z}$ be a representation of a partition $Z \in \S{P}$. Then there is a set $\S{F}$ with the following properties:
\begin{enumerate}
\item $\vec{Z} \in \S{F}$.
\item $\S{F}$ contains exactly one representation of each partition. 
\item The closure $\overline{\S{F}}$ is connected. 
\end{enumerate}
We call $\S{F}$ a fundamental region of $\vec{Z}$. The inverse $\phi = \pi^{-1}$ of the natural projection restricted to $\S{F}$ is measurable and induces a measure $q = \phi(Q)$ on $\S{F}$, which is the image measure of $Q$.

\begin{theorem}[Expected Partition Theorem]\label{theorem:EPT}
Let $Q$ be a probability measure on $\S{P}$.  Suppose that $M \in \S{P}$ is a local minimum of the expected Fr\'echet function $F_Q(Z)$. Then every representation $\vec{M} \in M$ is of the form 
\begin{align*}
\vec{M} = \int_{\S{F}} \vec{X}\, d q(\vec{X}), 
\end{align*}
where $\S{F}$ is a fundamental region of $\vec{M}$ and $q$ is the measure on $\S{F}$ induced by $Q$. 
\end{theorem}

\medskip

Note that the form of $\vec{M}$ is independent of the choice of fundamental region containing $\vec{M}$. Comparing both partition theorems, a representation $\vec{M}_{\!Q}$ of a local minimum of an expected Fr\'echet function $F_Q$ has a similar form as a representation $\vec{M}_{\!n}$ of a local minimum of a sample Fr\'echet Function $F_n$. Both, $\vec{M}_{\!Q}$ and $\vec{M}_{\!n}$, average over representations in optimal position.

\section{Applications of the Mean Partition Theorem}

This section presents examples that use the Mean Partition Theorem: (i) equivalence to multiple alignments, (ii) profiles and motifs, and (iii) cluster stability.

\subsection{Equivalence to Optimal Multiple Alignment}

\commentout{
Suppose that $\vec{Z}$ is a representation of some partition $Z$. A multiple alignment $\mathfrak{X}=\args{\vec{X}_{\!1}, \ldots, \vec{X}_{\!n}}$ is said to be in optimal position with $\vec{Z}$, if all representations $\vec{X}_{\!i}$ are in optimal position with $\vec{Z}$. 
}

We show that the problem of finding a mean partition is equivalent to the problem of finding an optimal multiple alignment of partitions.

\introskip

Let $\S{S}_n = \args{X_1, \ldots, X_n}$ be a sample of $n$ partitions $X_i \in \S{P}$. A multiple alignment of $\S{S}_n$ is an $n$-tuple $\mathfrak{X}= \args{\vec{X}_{\!1}, \ldots, \vec{X}_{\!n}}$
consisting of representations $\vec{X}_{\!i}\in X_{i}$. By
\[
\S{A}_n = \cbrace{\mathfrak{X} = \args{\vec{X}_{\!1}, \ldots, \vec{X}_{\!n}} \,:\, \vec{X}_{\!1} \in X_1, \ldots, \vec{X}_{\!n} \in X_n}
\]
we denote the set of all multiple alignments of $\S{S}_n$.

Next, we generalize the notion of optimal position to multiple alignments. Suppose that $\vec{Z}$ is a representation of some partition $Z$. A multiple alignment $\mathfrak{X}=\args{\vec{X}_{\!1}, \ldots, \vec{X}_{\!n}}$ is said to be in optimal position with $\vec{Z}$, if all representations $\vec{X}_{\!i}$ of $\mathfrak{X}$ are in optimal position with $\vec{Z}$. 

The mean of a multiple alignment $\mathfrak{X} = \args{\vec{X}_{\!1}, \ldots, \vec{X}_{\!n}}$ is denoted by
\[
\vec{M}_{\!\mathfrak{X}} = \frac{1}{n} \sum_{i=1}^n \vec{X}_{\!i}.
\] 
As shown in Lemma \ref{lemma:M_X-is-representation}, the mean $\vec{M}_{\!\mathfrak{X}}$ of a multiple alignment $\mathfrak{X}$ is an element of $\S{X}$ and therefore a representation of some partition $M_{\mathfrak{X}}$. It is important to note that a multiple alignment $\mathfrak{X}$ is not necessarily in optimal position with its mean $\vec{M}_{\!\mathfrak{X}}$. Thus, $\vec{M}_{\!\mathfrak{X}}$ does not necessarily satisfy the necessary conditions of optimality.

An optimal multiple alignment is a multiple alignment that minimizes the function 
\[
g_n\!\args{\mathfrak{X}} = \frac{1}{n^2}\sum_{i=1}^n \sum_{j=1}^n \normS{\vec{X}_{\!i} - \vec{X}_{\!j}}{^2}.
\]
The problem of finding an optimal multiple alignment is that of finding a multiple alignment with smallest average pairwise squared distances in $\S{X}$.  To show equivalence between  mean partitions and an optimal multiple alignments, we introduce the sets of minimizers of the respective functions $F_n$ and $g_n$:
\begin{align*}
\S{M}(F_n) &= \cbrace{M\in \S{P} \,:\, F_n(M) \leq F_n(Z)  \text{ for all } Z \in \S{P}}\\
\S{M}(g_n) &= \cbrace{\mathfrak{X} \in \S{A}_n \,:\, g_n(\mathfrak{X}) \leq g_n(\mathfrak{X}')  \text{ for all } \mathfrak{X}' \in \S{A}_n}
\end{align*}
For a given sample $\S{S}_n$, the set $\S{M}(F_n)$ is the mean partition set and $\S{M}(g_n)$ is the set of all optimal multiple alignments. The next result shows that any solution of $F_n$ is also a solution of $g_n$ and vice versa.
\begin{theorem}\label{theorem:equivalence:Fn-gn}
For any sample $\S{S}_n \in \S{P}^n$, the map
\[
\phi:\S{M}\!\args{g_n} \rightarrow \S{M}\!\args{F_n}, \quad \mathfrak{X} \mapsto \pi\!\args{\vec{M}_{\!\mathfrak{X}}}
\]
is surjective.
\end{theorem}

\medskip

Recall that $\pi:\S{X} \rightarrow \S{P}$ is the natural projection that sends matrices to partitions they represent. Theorem \ref{theorem:equivalence:Fn-gn} states that the mean $\vec{M}_{\!\mathfrak{X}}$ of an optimal multiple alignment $\mathfrak{X}$ is a representation of a mean partition $M_{\mathfrak{X}} = \pi\!\args{\vec{M}_{\!\mathfrak{X}}}$.

\commentout{
Theorem \ref{theorem:equivalence:Fn-gn} has the following implications:

\setcounter{part_counter}{0}
\begin{part}
Algorithms for multiple sequence alignment from Bioinformatics can be directly adopted for approximating a mean partition in consensus clustering. For this, we need to replace the sequence alignment score function by the partition distance $\delta$. 
\end{part}

\begin{part}
Inspired by sequence motifs, the mean partition can be used to identify highly conserved patterns. 
\end{part}

\begin{part}
The function $g_{n,k}$ is closely related to cluster stability. 
\end{part}
}

\subsection{Profiles and Motifs}

Profiles count the relative frequency with which a data point occurs in a cluster for a given multiple alignment. Motifs are subsets of data points for which there is a high consensus that they belong to the same cluster. Both concepts, profiles and motifs, can be derived from optimal multiple alignments using any dissimilarity function on partitions. When using the intrinsic metric $\delta$ induced by the Euclidean distance, equivalence of the mean partition and optimal multiple alignment provides direct and efficient ways to construct both, profiles and motifs. 

\introskip

We assume that $\args{\S{P}, \Delta}$ is a partition space endowed with a distance function of the general form
\[
\Delta(X, Y) = \min \cbrace{d(\vec{X}, \vec{Y}) \,:\, \vec{X} \in X, \vec{Y} \in Y}, 
\]
where $d(\vec{X}, \vec{Y})$ is a distance function on $\S{X}$. Let $\mathfrak{X}= \args{\vec{X}_{\!1}, \ldots, \vec{X}_{\!n}}$ be an optimal multiple alignment of  the sample $\S{S}_n = \args{X_1, \ldots, X_n}$, where optimality is with respect to the function
\[
\tilde{g}_n(\mathfrak{X}) = \sum_{i=1}^n \sum_{j=1}^n d\args{\vec{X}_i, \vec{X}_j}.
\] 
Observe that the problem of minimizing $\tilde{g}_n$ is not necessarily equivalent to the problem of minimizing the Fr\'echet function corresponding to the distance function $\Delta$.  

\medskip

A profile of $\S{S}_n$ is a matrix $\vec{P} = (p_{kj})$ with elements
\[
p_{kj} =  \frac{1}{n} \sum_{i=1}^n x_{kj}^{(i)},
\]
where $x_{kj}^{(i)}$ denotes the degree of membership of data point $j$ in cluster $c_k$ according to representation $\vec{X}_{\!i}$. Thus, $p_{kj}$ measures the average degree of membership of data point $z_j$ in cluster $c_k$.  High (low) average values $p_{kj}$ indicate a high (low) consensus among the sample partitions on assigning cluster label $c_k$ to data point $z_j$. In matrix notation, a profile $\vec{P}$ is identical to the mean $\vec{M}_{\mathfrak{X}}$ of the optimal multiple alignment $\mathfrak{X}$. But recall that $\vec{P}$ needs not to be a representation of a mean partition.

\medskip 

We can derive motifs from profiles. A motif is a subset $\S{C}$ of dataset $\S{Z}$ that is frequently occurring as a single cluster in the optimal multiple alignment $\mathfrak{X}$. Let $\tau \in (0.5, 1)$ denote the consensus threshold. We define the truncation of profile $\vec{P}$ as a  ($\ell \times m$)-matrix $\vec{P}^\tau = (p_{kj}^\tau)$ with elements
\[
p_{kj}^\tau = \begin{cases}
1 & \tau \leq  p_{kj} \\
0 & \text{otherwise}.
\end{cases}
\]
Since $\tau > 0.5$ each column of the truncation has at most one non-zero element. Therefore any data point of $\S{Z}$ is either member of exactly one cluster or belongs to no cluster. Thus, a truncation $\vec{P}^\tau$ represents a partition of a subset $\S{Z}^\tau$ of $\S{Z}$ defined by the non-zero columns of $\vec{P}^\tau$. The subsets 
\[
\S{C}_k^\tau = \cbrace{z_j \in \S{Z} \,:\, p_{kj}^\tau = 1} \subseteq \S{Z}^\tau, \quad k \in \cbrace{1, \ldots, \ell}
\]
are the motifs of sample $\S{S}_n$ corresponding to the multiple alignment $\mathfrak{X}$ and the consensus threshold $\tau$. The motifs form a partition of subset $\S{Z}^\tau$. Figure \ref{fig:motif} illustrates the concept of motif.

\begin{figure}[t]
\centering
\includegraphics[width=0.8\textwidth]{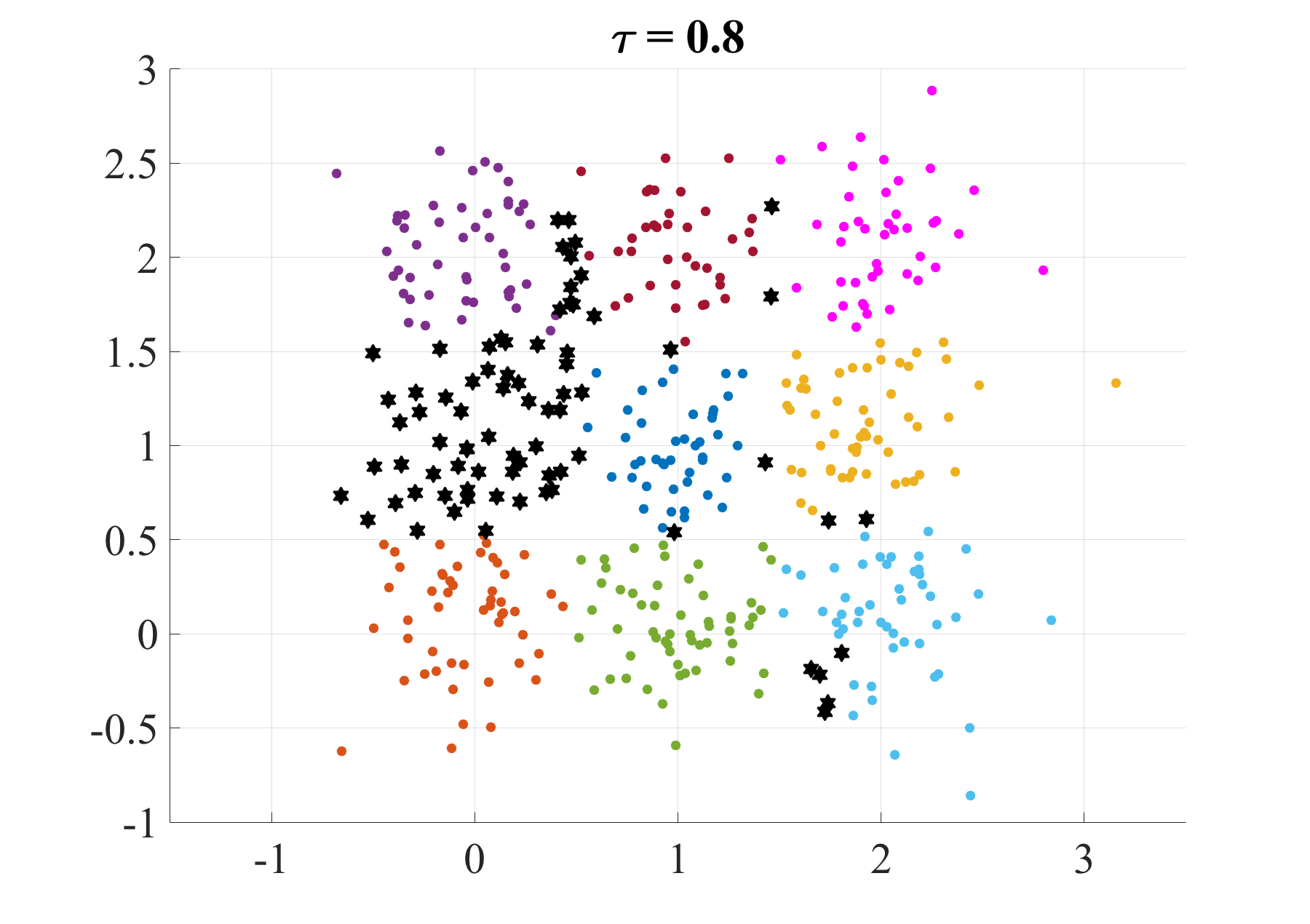}
\caption{Shown are motifs corresponding to consensus threshold $\tau = 0.8$. The motifs are derived from a sample of $100$ partitions obtained by k-means clustering ($k=9$) on a dataset drawn from a $3 \times 3$ grid of Gaussian distribution with identical variance. Data points of the same non-black color form a motif. Black stars refer to data points for which consensus among the partition is below $\tau = 0.8$. Note that consensus on the Gaussian component with center $(0,1)$ is below $\tau = 0.8$. For all other clusters consensus is high.}
\label{fig:motif}
\end{figure}

\commentout{
\begin{figure}[t]
\centering
\includegraphics[width=0.39\textwidth]{./distr_G9s03.png}
\hfill
\includegraphics[width=0.7\textwidth]{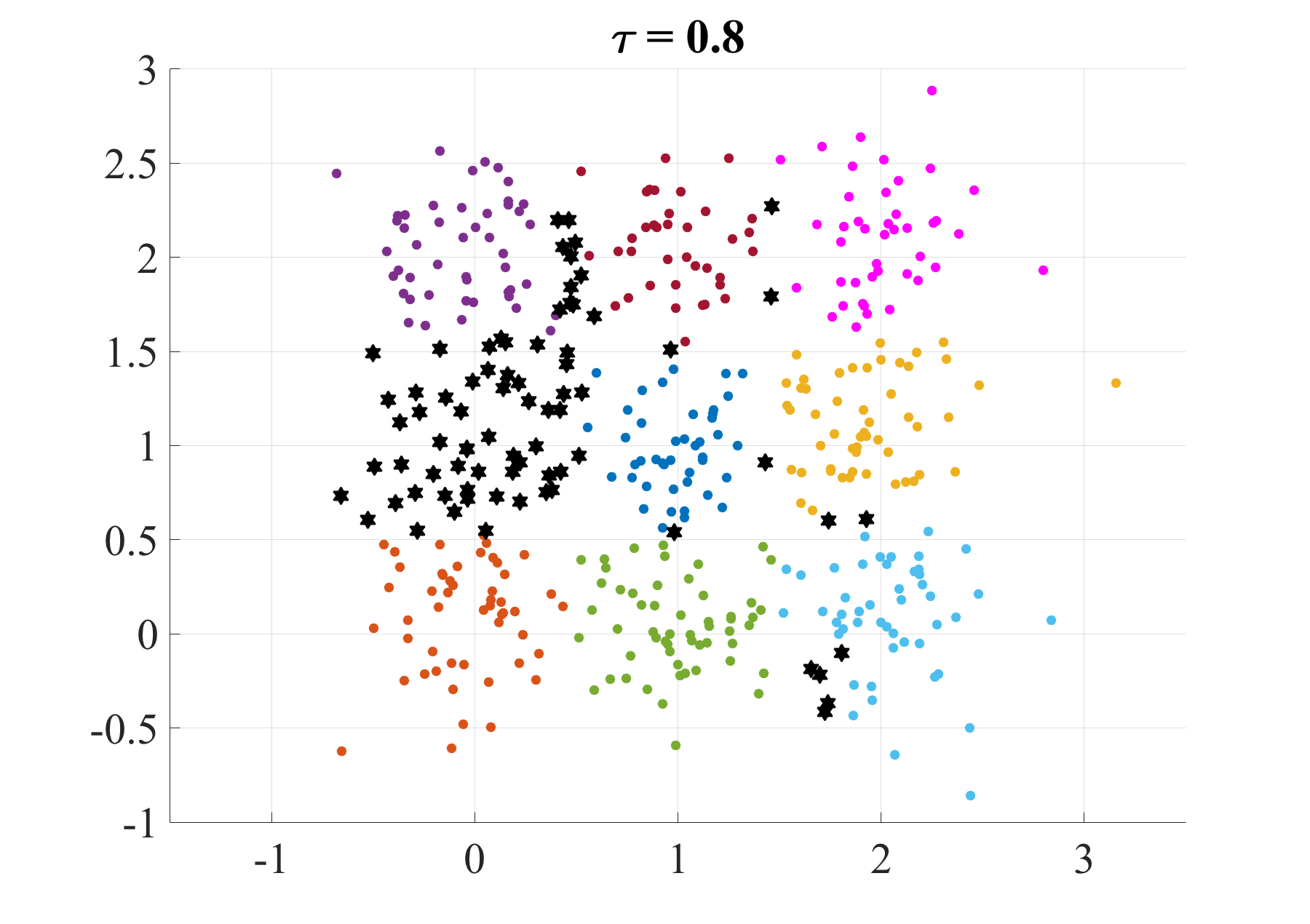}
\caption{Consensus profile (left) and motifs (right) derived from a sample of $100$ partitions of a dataset drawn from a $3 \times 3$ grid of Gaussian distribution with identical variance. The consensus profile shows how consensus of the partitions is distributed. A point $(\tau, P)$ on the consensus profile means that at least a fraction of $\tau$ partitions agree on $100\cdot P \%$ of all data points. Motifs corresponding to the consensus threshold $\tau = 0.8$ are clusters consisting of data points of the same non-black color. Black points are data points for which consensus is below $\tau = 0.8$.}
\label{fig:motif}
\end{figure}
}
\medskip

If the distance function $\Delta$ is the squared intrinsic metric  $\delta^2$, then the profile $\vec{P}$ is a representation of a mean partition. Hence, we can cast the problem of minimizing the function $g_n$ to the equivalent problem of minimizing the Fr\'echet function $F_n$ for which efficient approximate algorithms are available \cite{Dimitriadou2002,Gordon2001} that guarantee to converge to a local minimum of $F_n$.

\subsection{Cluster Stability}

This section applies the Mean Partition Theorem and its equivalence to the problem of multiple optimal alignment to cluster stability. For this, we follow a simplified setting for the sake of clarity.

\introskip

Choosing the number $\ell$ of clusters is a persisting model selection problem in clustering. One way to select $\ell$ is based on the concept of clustering stability. 
The intuitive idea behind clustering stability is that a clustering algorithm should produce similar partitions if repeatedly applied to slightly different datasets from the same underlying distribution. 

\medskip

Let $\S{P}_{k,m}$ be the set of partitions with $k$ clusters over $m$ data points from possibly different datasets. By $\delta_k$ we denote the metric on $\S{P}_{k,m}$ induced by the Euclidean norm. We assume that $\S{S}_{n, k} = \args{X_1, \ldots, X_n}$ is a sample of $n$ partitions $X_i \in \S{P}_{k, m}$. 

\medskip 

Following \cite{Luxburg2010}, model selection in clustering can be posed as the problem of minimizing some loss function
\[
\Gamma_n(k) : [k_{\min}, k_{\max}] \rightarrow \R, \quad k \mapsto \Gamma_{n, k}, 
\]
where $1 \leq k_{\min} \leq k \leq k_{\max} \leq m$. Then one simple option to choose the number $\ell$ of clusters is according to the rule 
\[
\ell = \arg\min_{k} \Gamma_{n, k}.
\]
A common and well established choice of loss function is cluster instability by optimal pairwise alignments \cite{Luxburg2010}
\[
G_{n, k} = \frac{1}{n^2} \sum_{i=1}^n \sum_{j = 1}^n\delta_k\!\argsS{X_i, X_j}{^2}.
\] 
Thus, we choose the number of clusters that gives the lowest average pairwise squared distances between partitions. In the following, we show that the mean partition problem of consensus clustering is related to the problem of cluster stability. 

\medskip

An alternative but related way to define cluster instability is by means of multiple instead of pairwise alignments. Let  $\mathfrak{X}_k= \args{\vec{X}_{\!1}, \ldots, \vec{X}_{\!n}}$ be an optimal multiple alignment of $\S{S}_{n,k}$. Then cluster instability based on optimal multiple alignment is defined by
\[
g_{n,k}\!\args{\mathfrak{X}_k} =  \frac{1}{n^2}\sum_{i=1}^n \sum_{j=1}^n \normS{\vec{X}_{\!i} - \vec{X}_{\!j}}{^2}.
\]
Observe that cluster instability by optimal pairwise and optimal multiple alignment are not equivalent, but related via the inequality 
\begin{align}\label{eq:G_kn < g_kn}
G_{n,k} \leq g_{n,k}\!\args{\mathfrak{X}_k}, 
\end{align}
because $\delta(X_i, X_j) \leq \norm{\vec{X}_{\!i} - \vec{X}_{\!j}}$ by definition. To illustrate the difference between $G_{n,k}$ and $g_{n,k}\!\args{\mathfrak{X}_k}$, we consider the following scenario:

\medskip

Suppose that $\mathfrak{X}_k = \args{\vec{X}, \vec{Y}, \vec{Z}}$ is an optimal multiple alignment of the three partitions $X$, $Y$, and $Z$, resp., such that 
\begin{align*}
\delta(X, Y) &= \norm{\vec{X}-\vec{Y}}\\
\delta(Y, Z) &= \norm{\vec{Y}-\vec{Z}}.
\end{align*}
Since being in optimal position is not a transitive property, it can happen that the representations $\vec{X}$ and $\vec{Z}$ are not in optimal position and therefore the inequality $\delta(X, Z) < \norm{\vec{X}-\vec{Z}}$ holds. 

\medskip

The implication is that instability $G_{n,k}$ by optimal pairwise alignments admits different representations of the same sample partition, whereas instability $g_{n,k}$ by optimal multiple alignment demands to use exactly one representation of each sample partition. Informally, optimal pairwise alignments as used in $G_{n,k}$ consider different interpretations of a clustering and optimal multiple alignments as used in $g_{n,k}$ consider a single interpretation of a clustering.

\medskip

Next, we present another path to cluster instability $g_{n,k}$ by multiple alignment. We can equivalently rewrite the instability score $G_{n,k}$ as 
\begin{align}\label{eq:Gnk-Fnk}
G_{n,k} = \frac{1}{n} \sum_{i=1}^n F_{n,k}\!\args{X_i},
\end{align}
where 
\[
F_{n, k}\!\args{X_i} = \frac{1}{n}\sum_{j = 1}^n\delta_k\!\argsS{X_i, X_j}{^2}
\]
is the average variation of sample $\S{S}_{n,k}$ with respect to sample partition $X_i$. This shows that cluster instability by pairwise alignments measures the average variation of sample $\S{S}_{n,k}$ with respect to all sample partitions. Thus, Eq.~\eqref{eq:Gnk-Fnk} links the instability score $G_{n,k}$ to consensus clustering and to Fr\'echet functions. 

Intuitively, we expect that the average pairwise distances $G_{n,k}$ between partitions and the average distance $F_{n,k}(M_k)$ to a mean partition $M_k$ are correlated. We have the relationship
\begin{align}\label{eq:F<I}
F_{n,k}\!\args{M_k} \leq G_{n, k},
\end{align}
where $M_k$ is a mean partition of sample $\S{S}_{n,k}$. These considerations suggest that the variation $F_{n,k}(M_k)$ can serve as an alternative score function for model selection that is related to cluster instability $G_{n,k}$. 

\medskip

\begin{figure}[t]
\centering
\includegraphics[width=0.49\textwidth]{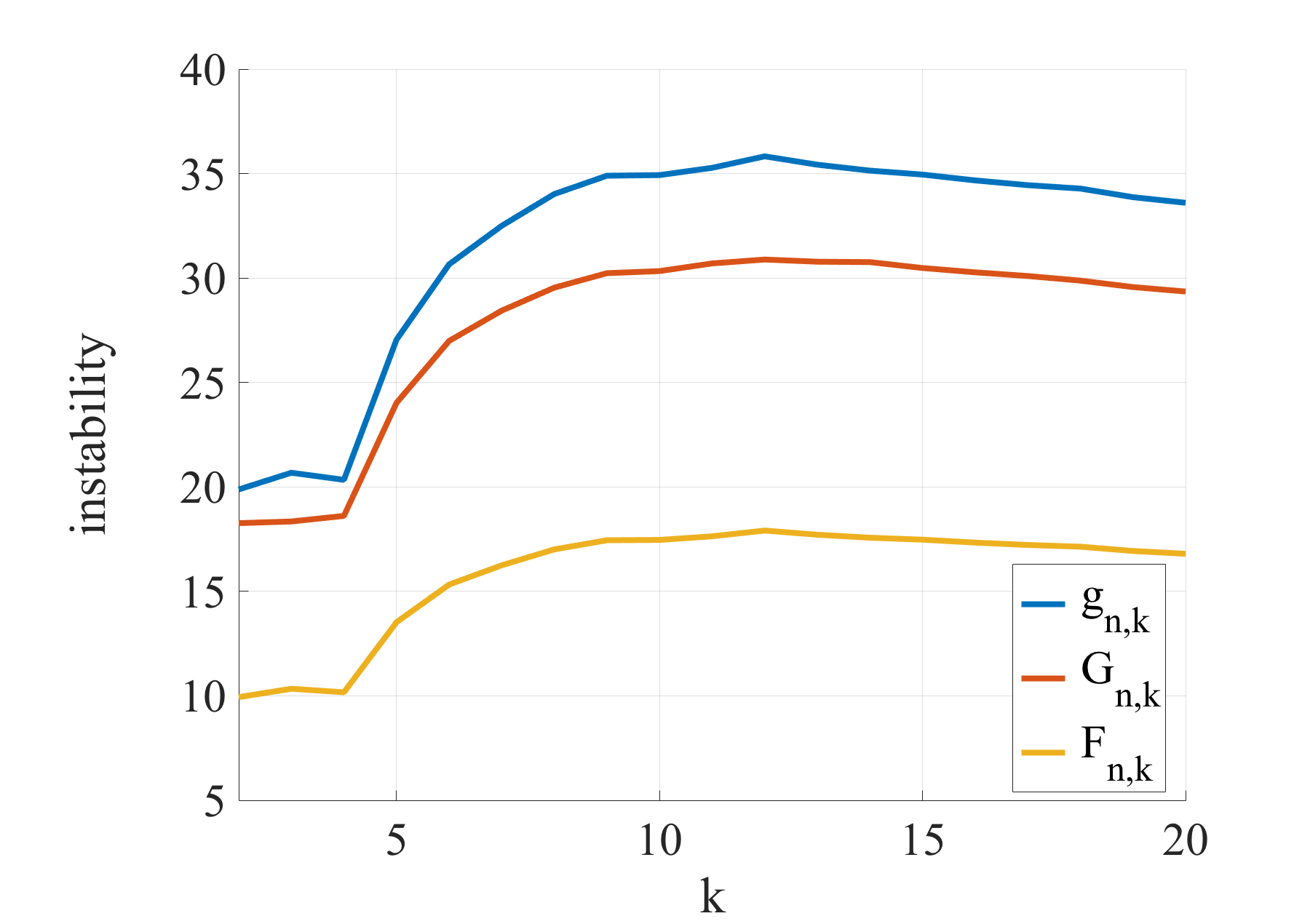}
\hfill
\includegraphics[width=0.49\textwidth]{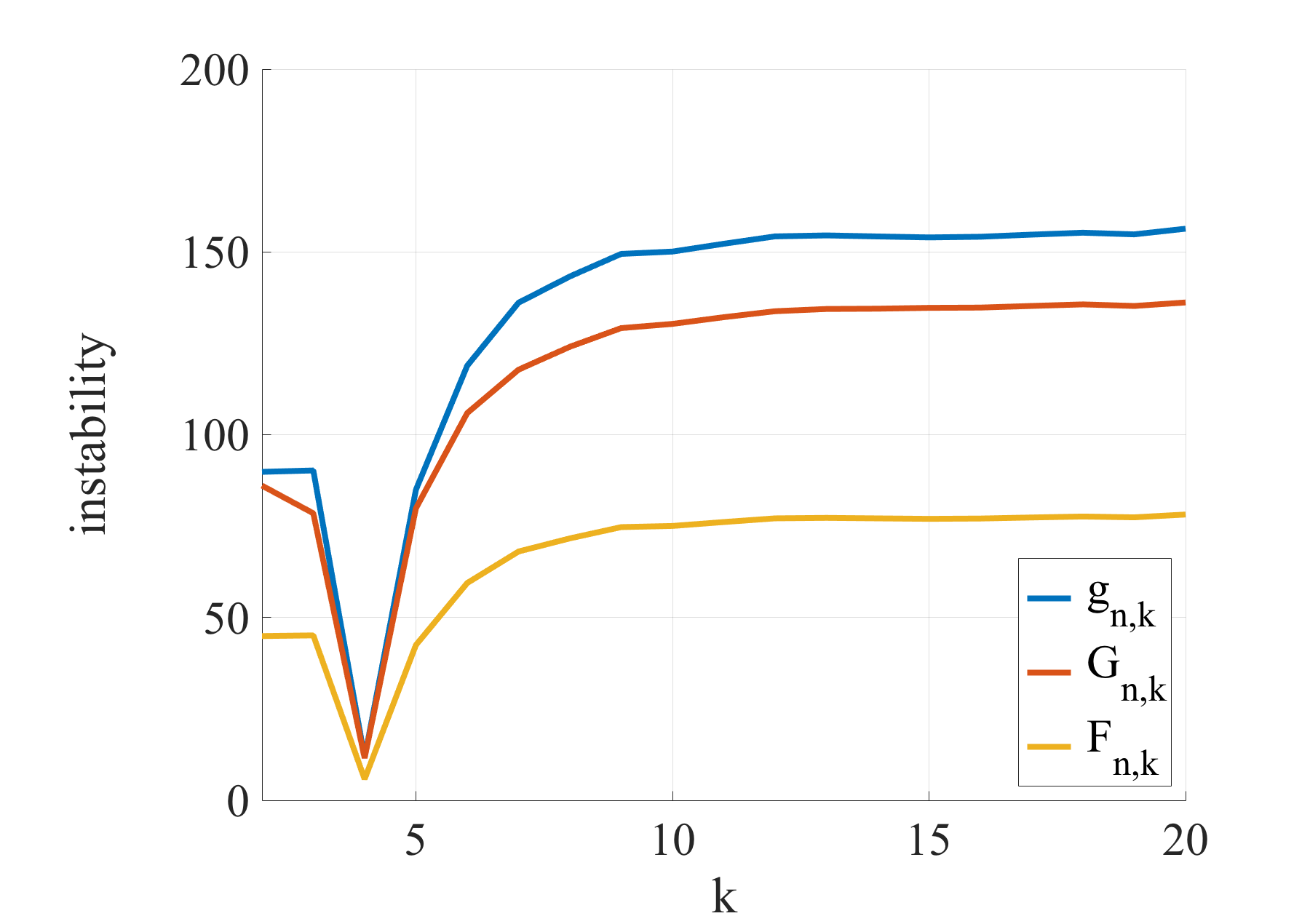}
\caption{Cluster instabilities $G_{n,k}$ (red), $g_{n,k}$ (blue), and $F_{n,k}$ (yellow) with $n = 100$ as a function of the number $k$ of clusters on uniform distribution (left) and on a $2 \times 2$ grid of  spherical Gaussian distribution with identical variance (right). Instabilities are averaged over $50$ randomly drawn datasets from each distribution.}
\label{fig:instabilities}
\end{figure}

From Theorem \ref{theorem:equivalence:Fn-gn} follows that cluster instability by variation $F_{n,k}(M_k)$ is equivalent to cluster instability $g_{n,k}\!\args{\mathfrak{X}_k}$ by multiple alignments. This observation has the following implications:

\setcounter{part_counter}{0}
\begin{part}
By combining inequalities \eqref{eq:G_kn < g_kn} and \eqref{eq:F<I} we obtain
\[
F_{n,k}\!\args{M_k} \leq G_{n, k} \leq g_{n,k}\!\args{\mathfrak{X}_k}. 
\]
Pairwise similar sample partitions imply low instability $G_{n,k}$ and low instability $G_{n,k}$ indicates high stability of the clustering. From the analysis on the uniqueness of the mean partition in \cite{Jain2016} follows that the difference $D = G_{n, k}-g_{n,k}\!\args{\mathfrak{X}_k}$ is more likely to diminish with increasing stability of the clustering. Conversely, the difference  $D$ is more likely to increase with decreasing stability of the clustering. These findings suggest that cluster instability $g_{n,k}\!\args{\mathfrak{X}_k}$ by multiple alignments more sharply emphasizes a stable clustering than clustering instability $G_{n,k}$ by pairwise alignments. Figure \ref{fig:instabilities} illustrates this behavior. 
\end{part}

\begin{part}
Determining cluster instability $g_{n,k}$ by multiple alignments is computationally intractable. Therefore, we need to resort to approximate solutions. Theorem \ref{theorem:equivalence:Fn-gn} justifies to cast the problem of approximating an optimal multiple alignment to the equivalent problem of determining a mean partition. For the latter problem, efficient algorithms for approximating the mean partition that converge to a local minimum of the Fr\'echet function are available \cite{Dimitriadou2002,Gordon2001}. Theorem \ref{theorem:MPT} gives us a multiple alignment $\mathfrak{X}'_k$ in optimal position to a representation of the local minimum. Then $g_{n,k}(\mathfrak{X}'_k)$ is an upper bound of the true cluster instability  $g_{n,k}$. 
\end{part}

\section{Conclusion}
The Mean Partition Theorem and the Expected Partition Theorem do not only provide necessary conditions of optimality for the Fr\'echet function based on the squared intrinsic metric, but also form the basis for interesting insights into consensus clustering. The problem of finding a mean partition is equivalent to the problem of finding an optimal multiple alignment. This result connects consensus clustering to computational biology. As an example, we constructed profiles and motifs for analyzing the result of a cluster ensemble. The third example 
relates consensus clustering to cluster stability. 

We hypothesize that the Mean Partition Theorem is pivotal for deriving further interesting results. Based on this hypothesis, we suggest four directions for future research: 
(1) Extending the finite sample results to asymptotic results by means of the Expected Mean Theorem and the consistency results presented in \cite{Jain2015c}; (2) further exploiting the Mean Partition Theorem for gaining new insight into consensus clustering; (3) generalizing the results to metrics other than the intrinsic metric; and (4) further exploiting ideas and techniques from computational biology for consensus clustering.

\begin{appendix}
\small
\section{Geometry of Partition Spaces}\label{app:sec:partition-spaces}

The proof of Theorem \ref{theorem:equivalence:Fn-gn} requires a suitable representation of partitions. We suggest to analyze partitions in a geometric framework by means of orbit spaces \cite{Jain2015c}. Orbit spaces are well explored, possess a rich geometrical structure and have a natural connection to Euclidean spaces \cite{Bredon1972,Jain2015,Ratcliffe2006}.

\subsection{Partition Spaces}

The group $\Pi = \Pi^\ell$ of all ($\ell \times \ell$)-of all ($\ell \times \ell$)-permutation matrices acts on $\S{X}$ by matrix multiplication, that is
\[
\cdot : \Pi \times \S{X} \rightarrow \S{X}, \quad (\vec{P}, \vec{X}) \mapsto \vec{PX}.
\]
The orbit of $\vec{X} \in \S{X}$ is the set $\bracket{\vec{X}} = \cbrace{\vec{PX} \,:\, \vec{P} \in \Pi}$. The orbit space of partitions is the quotient space $\S{X}/\Pi = \cbrace{\bracket{\vec{X}} \,:\, \vec{X} \in \S{X}}$ obtained by the action of the permutation group $\Pi$ on the set $\S{X}$. We write $X \in \S{P}$ to denote an orbit $[\vec{X}] \in \S{X}/\Pi$.

The partition space $\S{P}$ is endowed with the intrinsic metric $\delta$ induced by the Euclidean distance on $\S{X}$. For our purposes, it is more convenient to approach $\delta$ in a slightly different way. Let 
\[
\inner{\vec{X}, \vec{Y}}= \sum_{k = 1}^\ell \sum_{j = 1}^m x_{kj} y_{kj}
\]
denote the inner product of $\vec{X}, \vec{Y} \in \S{X}$. The inner product induces a well-define length on partitions:
\[
N(X) = \max \cbrace{\sqrt{\inner{\vec{X}, \vec{X}'}} \,:\, \vec{X}, \vec{X}' \in X}.
\]
The next result shows that the length of a partition can be computed in a straightforward way. 
\begin{proposition}\label{prop:N(X)}
Let $X \in \S{P}$ be a partition. Then 
\[
N(X) = \sqrt{\inner{\vec{X}, \vec{X}}}
\]
for all representations $\vec{X} \in X$. 
\end{proposition}
\proof
See \cite{Jain2015}, Prop.~3.17.
\qed

\medskip

\noindent
The squared intrinsic metric takes the equivalent form \cite{Jain2015}
\[
\delta(X, Y)^2 = \min \cbrace{N(X)^2 - 2 \inner{\vec{X}, \vec{Y}} + N(Y)^2 \,:\, \vec{X} \in X, \vec{Y} \in Y}.
\]

\subsection{Notations}

We use the following notations:  By $\overline{\S{U}}$ we denote the closure of a subset $\S{U} \subseteq \S{X}$, by $\partial \S{U}$ the boundary of $\S{U}$, and by $\S{U}^\circ$ the open subset $\overline{\S{U}} \setminus \partial \S{U}$. The action of permutation $\vec{P} \in \Pi$ on the subset $\S{U}\subseteq \S{X}$ is the set defined by $\vec{P}\,\S{U} = \cbrace{\vec{PX} \, :\, \vec{X} \in \S{U}}$. The action of a subset $\Phi \subseteq \Pi$ on $\S{U}$ is defined by $\Phi \,\S{U} = \cbrace{\vec{PX} \,:\, \vec{P} \in \Phi, \vec{X} \in \S{U}}$. By $\Pi^* = \Pi \setminus \cbrace{\vec{I}}$ we denote the subset of ($\ell \times \ell$)-permutation matrices without identity matrix $\vec{I}$.

\subsection{Voronoi Cells}

Let $\vec{Z}$ be a representation of partition $\vec{Z}$. The set 
\[
\S{V}_{\vec{Z}} = \cbrace{\vec{X} \in \S{X} \,:\, \norm{\vec{X}-\vec{Z}} \leq  \norm{\vec{X}-\vec{PZ}} \text{ for all } \vec{P} \in \Pi}
\]
is the Voronoi cell of $\vec{Z}$. The form of a Voronoi cell of $\vec{Z}$ depends on the form of the stabilizer of $\vec{Z}$. The stabilizer subgroup of $\Pi$ with respect to $\vec{Z}$ is defined by
\[
\Pi_{\vec{Z}} = \cbrace{\vec{P} \in \Pi \,:\, \vec{PZ} = \vec{Z}}.
\] 
We say, partition $Z$ is asymmetric if the stabilizer $\Pi_{\vec{Z}}$ of a representation $\vec{Z} \in Z$ is trivial, that is $\Pi_{\vec{Z}}  = \cbrace{\vec{I}}$. Otherwise, we call the partition $Z$ symmetric. 

The definition of asymmetry is well-defined, that is independent of the choice of representation, because stabilizers of elements in the same orbit are conjugate to each other. In more detail, let $\vec{Z}$ and $\vec{Z}'$ are representations of $Z$ with $\vec{Z}' = \vec{PZ}$ for some $\vec{P} \in \Pi$. Then we have $\Pi_{\vec{Z}'} = \vec{P}^{-1}\Pi_{\vec{Z}} \vec{P}$ by conjugacy of the stabilizers. Thus, $Z$ is asymmetric if and only if there is a representation $\vec{Z}$ with trivial stabilizer. Then conjugacy implies that the stabilizers of all representations $\vec{Z}'$ of $Z$ are trivial.

For any $\vec{P} \in \Pi$ the set $\vec{P} \Pi_{\vec{Z}} = \cbrace{\vec{PQ} \:,\: \vec{Q} \in \Pi_{\vec{Z}}}$ is a left coset of $\Pi_{\vec{Z}}$ in $\Pi$. The quotient set $\Pi/\Pi_{\vec{Z}} = \cbrace{\vec{P}\Pi_{\vec{Z}}\,:\, \vec{P} \in \Pi}$ consists of all left cosets of $\Pi_{\vec{Z}}$ in $\Pi$. We briefly denote left cosets $\vec{P}\Pi_{\vec{Z}}$ by $\bracket{\vec{P}}$.

In the following, we show a few auxiliary results. 

\begin{lemma}\label{lemma:asymmetry-is-generic} 
Almost all partitions are asymmetric. 
\end{lemma}

\proof \cite{Jain2016}, Prop.~3.3. \qed

\begin{lemma}\label{lemma:reps-in-V_Z}
Let $\vec{Z}$ be a representation of partition $\vec{Z}$. Suppose that $\vec{X} \in \S{V}_{\vec{Z}}$. Then $\vec{PX} \in \S{V}_{\vec{Z}}$ for all $\vec{P} \in \Pi_{\vec{Z}}$.
\end{lemma}

\proof
From $\vec{X} \in \S{V}_{\vec{Z}}$ follows $\norm{\vec{X}-\vec{Z}} \leq  \norm{\vec{X}-\vec{QZ}}$ for all $\vec{Q} \in \Pi$. Since $\Pi$ acts isometrically on $\S{X}$, we have
$\norm{\vec{X}-\vec{Z}} =  \norm{\vec{PX}-\vec{PZ}}$. From $\vec{P} \in \Pi_{\vec{Z}}$ follows $\vec{Z} =\vec{PZ}$ and therefore $\norm{\vec{PX}-\vec{PZ}} = \norm{\vec{PX}-\vec{Z}}$. Combining the inequalities and equations yields  
\[
\norm{\vec{X}-\vec{Z}} = \norm{\vec{PX}-\vec{Z}}  = \norm{\vec{PX}-\vec{PZ}} \leq  \norm{\vec{PX}-\vec{QPZ}} =  \norm{\vec{PX}-\vec{QZ}}.
\]
This shows the assertion.
\qed

\medskip

\begin{lemma}\label{lemma:PV_Z= V_PZ}
Let $\vec{Z}$ be a representation of partition $\vec{Z}$. Then 
\[
\S{V}_{\vec{PZ}} = \vec{P} \,\S{V}_{\vec{Z}} = \bracket{\vec{P}} \S{V}_{\vec{Z}}
\]
for all $\vec{P} \in \Pi$.
\end{lemma}

\proof \ 
\setcounter{part_counter}{0}
\begin{part}
For any $\vec{X} \in \S{V}_{\vec{PZ}}$ we have $\norm{\vec{X} - \vec{PZ}} \leq \norm{\vec{X} - \vec{QPZ}}$ for all $\vec{Q} \in \Pi$. Since $\Pi$ acts isometrically on $\S{X}$, we find that 
\[
\norm{\vec{P}^{-1}\vec{X} - \vec{Z}} \leq \norm{\vec{P}^{-1}\vec{X} - \vec{P}^{-1}\vec{QPZ}}
\] 
for all $\vec{Q} \in \Pi$. From $\vec{P}^{-1}\Pi\vec{P} = \Pi$ follows that $\vec{P}^{-1}\vec{X} \in \S{V}_{\vec{Z}}$ and therefore $\vec{X} \in \vec{P}\S{V}_{\vec{Z}}$. 
\end{part}

\begin{part}
For any $\vec{Q} \in \Pi_{\vec{Z}}$ we have  $\vec{Z} = \vec{QZ}$. Hence,  for any element $\vec{PQ}$ of the left coset $\vec{P}\Pi_{\vec{Z}}$ we have $\vec{X} \in \S{V}_{\vec{PZ}} = \S{V}_{\vec{PQZ}} = \vec{PQ}\S{V}_{\vec{Z}}$. This shows $\vec{X} \in \bracket{\vec{P}}\S{V}_{\vec{Z}}$.
\qed
\end{part}

\medskip

\begin{lemma}\label{lemma:decomposition-of-V_z}
Let $\vec{Z}$ be a representation of partition $\vec{Z}$. Then 
\begin{enumerate}
\item $\displaystyle \S{X}  = \bigcup_{\bracket{\vec{P}} \in \Pi/\Pi_{\vec{Z}}} \bracket{\vec{P}}\S{V}_{\vec{Z}}$.
\item $\bracket{\vec{P}} \S{V}_{\vec{Z}}^\circ \cap \S{V}_{\vec{Z}}^\circ = \emptyset$ for all $\bracket{\vec{P}} \in \Pi^*/\Pi_{\vec{Z}}$.
\end{enumerate}
\end{lemma}

\proof \ 
\setcounter{part_counter}{0}
\begin{part}
Since $\Pi$ is a group action on $\S{X}$, it is sufficient to show the $'\!\!\!\subseteq'$-direction. Let $\vec{X} \in \S{X}$ be a representation of partition $X \in \S{P}$.  Then there is a representation $\vec{X}'$ of $X$ and a $\vec{P} \in \Pi$ such that $\vec{X}' \in \S{V}_{\vec{Z}}$ and $\vec{X} = \vec{PX}'$. This shows that $\vec{X} \in \vec{P}\S{V}_{\vec{Z}} =  \bracket{\vec{P}}\S{V}_{\vec{Z}}$ by Lemma \ref{lemma:PV_Z= V_PZ}. This proves the first assertion.
\end{part}

\begin{part}
Let $\vec{P} \in \Pi^*$ such that $\bracket{\vec{P}} \neq \bracket{\vec{I}}$. Then we have $\vec{PZ} \neq \vec{Z}$. Let $\vec{X} \in \S{V}_{\vec{Z}}^\circ$ be an element of the interior of  $\S{V}_{\vec{Z}}$. Then we have
\[
\norm{\vec{X} - \vec{Z}} < \norm{\vec{X} - \vec{P}\vec{Z}}.
\]
This shows that $\vec{X} \notin \S{V}_{\vec{PZ}} = \bracket{\vec{P}} \S{V}_{\vec{Z}}$. In a similar way we show that $\vec{X} \in \bracket{\vec{P}} \S{V}_{\vec{Z}}^\circ$ with $\bracket{\vec{P}} \neq \bracket{\vec{I}}$ is not contained in $\S{V}_{\vec{Z}}$, which proves the second assertion.
\qed
\end{part}

\subsection{Dirichlet Fundamental Domains}

A subset $\S{F}$ of $\S{X}$ is a fundamental set for $\Pi$ if and only if $\S{F}$ contains exactly one representation $\vec{X}$ from each orbit $\bracket{\vec{X}} \in \S{X}/\Pi$. 
A fundamental domain of $\Pi$ in $\S{X}$ is a closed connected set $\S{F} \subseteq \S{X}$ that satisfies 
\begin{enumerate}
\item $\displaystyle\S{X} = \bigcup_{\vec{P} \in \Pi} \vec{P}\S{F}$
\item $\vec{P} \S{F}^\circ \cap \S{F}^\circ = \emptyset$ for all $\vec{P} \in \Pi^*$.
\end{enumerate}

\begin{proposition}
Let $\vec{Z}$ be a representation of an asymmetric partition $Z \in \S{P}$. Then 
\[
\S{D}_{\vec{Z}} = \cbrace{\vec{X} \in \S{X} \,:\, \norm{\vec{X} - \vec{Z}} \leq \norm{\vec{X} - \vec{PZ}} \text{ for all }\vec{P} \in \Pi}
\]
is a fundamental domain, called Dirichlet fundamental domain of $\vec{Z}$. 
\end{proposition}

\proof
 \cite{Ratcliffe2006}, Theorem 6.6.13. 
 \qed

\medskip

We extend the notion of Dirichlet fundamental domain to arbitrary partitions. For this, we need the following result.

\begin{proposition}\label{prop:prop:Voronoi-contains-Dirichlet:Fj=QFi}
Let $\vec{Z}$ be a representation of a partition $Z \in \S{P}$ with stabilizer $\Pi_{\vec{Z}}$. Then there is a fundamental domain $\S{F}$ satisfying
\[
\displaystyle\S{V}_{\vec{Z}} = \bigcup_{\vec{P} \in \Pi_{\vec{Z}}} \vec{P}\S{F}.
\]
\end{proposition}

\proof
\setcounter{part_counter}{0}
\begin{part}
Let $\Pi_{\vec{Z}} = \cbrace{\vec{P}_{\!1}, \ldots, \vec{P}_{\!q}}$ be the stabilizer subgroup of $\Pi$ with respect to $\vec{Z}$. Suppose that $\vec{X}_{\!0} \in \S{V}_{\vec{Z}}$ is a representation of an asymmetric  partition $X_0$. Such an element $\vec{X}_{\!0}$ exists, because the natural projection $\pi: \S{V}_{\vec{Z}} \rightarrow \S{P}$ is surjective and by Lemma \ref{lemma:asymmetry-is-generic}. Let 
\[
\S{F}_{0} = \cbrace{\vec{X} \in \S{V}_{\vec{Z}} \,:\, \pi\!\args{\vec{X}} = X_0}
\]
denote the fiber over $X_0$ restricted to $\S{V}_{\vec{Z}}$. Since $X_0$ is asymmetric, the fiber over $X_0$ is of the form 
\[
\S{F}_{0} = \cbrace{\vec{PX}_{\!0} \,:\, \vec{P} \in \Pi_{\vec{Z}}}
\]
and has exactly $q$ elements. For every $i \in \cbrace{1, \ldots, q}$, we define the set
\[
\S{F}_i = \cbrace{\vec{X} \in \S{V}_{\vec{Z}} \,:\, \norm{\vec{X} - \vec{P}_{\!i}\vec{X}_{\!0}} \leq \norm{\vec{X} - \vec{P}_{\!j}\vec{X}_{\!0}} \text{ for all } j \in \cbrace{1, \ldots, q}}.
\]
The sets $\S{F}_i$ form a Voronoi tesselation of $\S{V}_{\vec{Z}}$ with centers $\vec{P}_{\!i}\vec{X}_{\!0}$ and are therefore closed, convex and connected subsets of $\S{V}_{\vec{Z}}$ satisfying
\[
\S{V}_{\vec{Z}} = \bigcup_{i=1}^q \S{F}_i.
\]
and $\S{F}_i^\circ \cap \S{F}_j^\circ = \emptyset$ for all $1 \leq i < j \leq q$.
\end{part}

\begin{part}
We show that $\S{F}_j = \vec{P}_{\!j} \vec{P}_{\!i}^{-1}\S{F}_i$ for all $i, j \in \cbrace{1, \ldots, q}$. We set $\vec{Q}_{ij} = \vec{P}_{\!j} \vec{P}_{\!i}^{-1}$ and first show $\vec{Q}_{ij}\S{F}_i \subseteq \S{F}_j$. Let $\vec{X} \in \S{F}_i$. Since $\Pi$ acts isometrically on $\S{X}$, we have
\[
 \norm{\vec{X} - \vec{P}_{\!i}\vec{X}_{\!0}} = \norm{\vec{P}_{\!j} \vec{P}_{\!i}^{-1} \vec{X} - \vec{P}_{\!j}\vec{X}_{\!0}} = \norm{\vec{Q}_{ij} \vec{X} - \vec{P}_{\!j}\vec{X}_{\!0}}.
\]
Since a stabilizer is a subgroup, we finde that $\vec{Q}_{ij} \in \Pi_{\vec{Z}}$. Then from Lemma \ref{lemma:reps-in-V_Z} follows that $\vec{Q}_{ij}\vec{X} \in \S{V}_{\vec{Z}}$. This together with isometry of $\Pi$ implies that $\vec{Q}_{ij} \vec{X} \in \S{F}_j$. Thus, we have  $\vec{Q}_{ij}\S{F}_i \subseteq \S{F}_j$. Now we assume that $\vec{X} \in \S{F}_j$. By isometry we have 
\[
 \norm{\vec{X} - \vec{P}_{\!j}\vec{X}_{\!0}} = \norm{\vec{P}_{\!i}\vec{P}_{\!j}^{-1} \vec{X} - \vec{P}_{\!i}\vec{X}_{\!0}} = \norm{\vec{Q}_{ij}^{-1} \vec{X} - \vec{P}_{\!i}\vec{X}_{\!0}}.
 \]
 The inverse $\vec{Q}_{ij}^{-1}$ is an element of the stabilizer $\Pi_{\vec{Z}}$, because $\vec{Q}_{ij} \in \Pi_{\vec{Z}}$ and the stabilizer is a group. From Lemma \ref{lemma:reps-in-V_Z} follows that $\vec{Q}_{ij}^{-1}\vec{X} \in \S{V}_{\vec{Z}}$. Hence,  we find that $\vec{Q}_{ij}^{-1}\vec{X} \in \S{F}_i$. This shows that $\vec{Q}_{ij}\vec{Q}_{ij}^{-1}\vec{X} \in \vec{Q}_{ij}\S{F}_i$ giving $\S{F}_j \subseteq \vec{Q}_{ij}\S{F}_i$. 
\end{part}

\begin{part}
Let $\S{F} = \S{F}_i$ for some $i \in \cbrace{1, \ldots, q}$. It remains to show that $\S{F}$ is a fundamental domain. From the first two parts of this proof follows that $\S{F}$ is closed and connected (as a convex set) such that
\[
\S{V}_{\vec{Z}} = \bigcup_{\vec{P} \in \Pi_{\vec{Z}}} \vec{P}\S{F}.
\]
From Lemma \ref{lemma:decomposition-of-V_z}(1) follows
\[
\S{X} = \bigcup_{\vec{Q} \in \Pi} \vec{Q} \S{V}_{\vec{Z}} = \bigcup_{\vec{Q} \in \Pi} \vec{Q} \bigcup_{\vec{P} \in \Pi_{\vec{Z}}} \vec{P}\S{F} = \bigcup_{\vec{Q} \in \Pi}\bigcup_{\vec{P} \in \Pi_{\vec{Z}}} \vec{QP}\S{F} = \bigcup_{\vec{Q} \in \Pi}\vec{Q}\S{F}.
\]
From Part 1 of this proofs follows $\S{F}^\circ \cap \vec{P}\S{F}^\circ = \emptyset$ for all $\vec{P} \in \Pi_{\vec{Z}}$. Applying Lemma \ref{lemma:decomposition-of-V_z}(2) extends this property over the entire set $\Pi$. 
\qed
\end{part}

\medskip

We call the fundamental domain $\S{F}$ in Prop.~\ref{prop:prop:Voronoi-contains-Dirichlet:Fj=QFi} a Dirichlet fundamental domain of $\vec{Z}$. In contrast to representations of asymmetric partitions, Dirichlet fundamental domains of representations of symmetric partitions are not uniquely determined.

\subsection{Cross Sections}

Suppose that $\S{D}_{\vec{Z}}$ is the Dirichlet fundamental domain of representation $\vec{Z}$ of an asymmetric partition $Z \in \S{P}$. A map $\mu:\S{P} \rightarrow \S{D}_{\vec{Z}}$ is a cross section into $\S{D}_{\vec{Z}}$, if $\pi(\mu(X)) = X$ for all partitions $X \in \S{P}$. Cross sections exist, because there is a fundamental set $\S{F}$ such that $\S{D}_{\vec{Z}}^\circ \subset \S{F} \subset \S{D}_{\vec{Z}}$ by \cite{Ratcliffe2006}, Theorem 6.6.11.

\begin{proposition}
Let $\mu:\S{P} \rightarrow \S{D}_{\vec{Z}}$ be a cross section into a Dirichlet fundamental domain $\S{D}_{\vec{Z}}$ of representation $\vec{Z}$ of partition $Z \in \S{P}$. Then the following properties hold:
\begin{enumerate}
\item $\mu$ is injective.
\item $\mu(\S{P})$ is a fundamental set. 
\item $\mu$ is a measurable mapping.
\end{enumerate}
\end{proposition}

\noindent
\proof \  
\setcounter{part_counter}{0}
\begin{part}
Injectivity of $\mu$ directly follows from the property $\pi\circ \mu = \id$.
\end{part}

\begin{part}
Again from $\pi\circ \mu = \id$ follows that $\mu$ maps partitions to representations.  Since $\mu$ is injective, the image $\mu(\S{P})$ contains exactly one representation of each partition. Hence, $\mu(\S{P})$ is a fundamental set.
\end{part}

\begin{part}
Finally, $\mu$ is measurable, because $\mu^{-1} = \pi$ and $\pi$ is continuous and open. 
\qed
\end{part}

\medskip 

Let $\args{\S{P}, \S{B}, Q}$ be a measurable space. A cross section $\mu:\S{P} \rightarrow \S{D}_{\vec{Z}}$ is a measurable map that gives rise to a measurable space $\args{\S{D}_{\vec{Z}}, \S{B}_\mu, q}$. 

\section{Proofs}

\subsection{Proof of Theorem \ref{theorem:MPT}}

\setcounter{part_counter}{0}
\begin{part}
The pull-back of $F_n$ is a function defined by
\[
f_n: \S{X} \rightarrow \R, \quad \vec{Z} \mapsto \frac{1}{n} \sum_{i=1}^n \min_{\vec{X}'_{\!i} \in X_i} \normS{\vec{X}'_{\!i} - \vec{Z}}{^2}.
\]
Observe that $f_n = F_n \circ \pi$. Let $\vec{M}$ be a local minimum of $f_n$. There is a multiple alignment $\mathfrak{X} = \args{\vec{X}_{\!1}, \ldots, \vec{X}_{\!n}}$ in optimal position with $\vec{M}$ such that 
\[
f_n\args{\vec{M}} =  \frac{1}{n} \sum_{i=1}^n \norm{\vec{X}_{\!i} - \vec{M}}{^2}.
\]
\end{part}
\begin{part}
The function
\[
f_{\mathfrak{X}}(\vec{Z}) = \frac{1}{n} \sum_{i=1}^n \normS{\vec{X}_{\!i} - \vec{Z}}{^2}.
\]
is differentiable and convex. By taking the gradient of $f_{\mathfrak{X}}$, equating to zero and solving the equation, we obtain the mean $\vec{M}_{\!\mathfrak{X}}$ of $\mathfrak{X}$ as unique minimum of $f_{\mathfrak{X}}$. Moreover, we have  
\begin{align}
\label{eq:proof:theorem:MPT:fn=fx}
 f_n(\vec{M}) &= f_{\mathfrak{X}}(\vec{M})  \geq f_{\mathfrak{X}}\args{\vec{M}_{\!\mathfrak{X}}} \\
 \label{eq:proof:theorem:MPT:fn<fx}
 f_n(\vec{Z}) & \leq f_{\mathfrak{X}}(\vec{Z}) 
\end{align}
for all $\vec{Z} \in \S{X}$ by construction. 
\end{part}
\begin{part}
We distinguish between two cases: 
\begin{enumerate}
\item $f_{\mathfrak{X}}(\vec{M})  = f_{\mathfrak{X}}\args{\vec{M}_{\!\mathfrak{X}}}$: This directly implies the assertion $\vec{M} = \vec{M}_{\!\mathfrak{X}}$, because $f_{\mathfrak{X}}$ is convex and has a unique minimum. 
\item $f_{\mathfrak{X}}(\vec{M})  > f_{\mathfrak{X}}\args{\vec{M}_{\!\mathfrak{X}}}$: We show that this case contradicts our assumption that $\vec{M}$ is a local minimum. Let $\S{B}_\varepsilon = \S{B}(\vec{M}, \varepsilon)$ denote the ball with center $\vec{M}$ and radius $\varepsilon > 0$. Then for every $\varepsilon > 0$ there is a representation $\vec{Z} \in \S{B}$ satisfying
\[
f_{\mathfrak{X}}(\vec{Z}) < f_{\mathfrak{X}}(\vec{M}) 
\]
because $f_{\mathfrak{X}}$ is convex and $\vec{M}$ is not a minimum of $f_{\mathfrak{X}}$. From Eq.~\eqref{eq:proof:theorem:MPT:fn=fx} and \eqref{eq:proof:theorem:MPT:fn<fx} follows that 
\[
 f_n(\vec{Z}) \leq f_{\mathfrak{X}}(\vec{Z}) < f_{\mathfrak{X}}(\vec{M}) = f_n(\vec{M}).
\]
Since $\varepsilon$ can be arbitrarily small, we find that $\vec{M}$ is not a local minimum, which contradicts our assumption. Hence, the first case applies. 
\end{enumerate}
\end{part}

\subsection{Proof of Theorem \ref{theorem:EPT}}

The proof follows a similar line as the proof of Theorem \ref{theorem:MPT}. 
\setcounter{part_counter}{0}
\begin{part}
We define the pull-back of $F_Q$ by
\[
f_Q: \S{X} \rightarrow \R, \quad \vec{Z} \mapsto  \int_{\S{D}_{\vec{Z}}} \normS{\vec{X}-\vec{Z}}{^2} d q(\vec{X}).
\]
where $q$ is the image measure of measure $Q$ under cross section $\mu$. We omit the dependence of $q$ and $\mu$ from the particular Dirichlet fundamental domain. 
The pull-back satisfies $f_Q = F_Q \circ \pi$. Let $\vec{M}$ be a local minimum of $f_Q$. 
\end{part}

\begin{part}
We define the function
\[
f_{\vec{M}}(\vec{Z}) = \int_{\S{D}_{\vec{M}}} \normS{\vec{X}-\vec{Z}}{^2} d q(\vec{X}).
\]
The function $f_{\vec{M}}(\vec{Z})$ is differentiable and convex. By taking the gradient of $f_{\vec{M}}$, equating to zero and solving the equation, we obtain 
\[
\vec{M}' = \int_{\S{D}_{\vec{M}}} \vec{X} d q(\vec{X}). 
\]
as unique minimum of $f_{\vec{M}}$. 
In line with Eq.~\eqref{eq:proof:theorem:MPT:fn=fx} and \eqref{eq:proof:theorem:MPT:fn<fx} of Theorem \ref{theorem:MPT} we have 
\begin{align*}
 f_Q(\vec{M}) &= f_{\vec{M}}(\vec{M})  \geq f_{\vec{M}}\args{\vec{M}'} \\
 f_Q(\vec{Z}) & \leq f_{\vec{M}}(\vec{Z}) 
\end{align*}
\end{part}

\begin{part}
We have $\vec{M} = \vec{M}'$ as in Part 3 of Theorem \ref{theorem:MPT}. 
\end{part}

\subsection{Lemma \ref{lemma:M_X-is-representation}}
\begin{lemma}\label{lemma:M_X-is-representation}
The mean $\vec{M}_{\!\mathfrak{X}}$ of a multiple alignment $\mathfrak{X}$ represents a partition.
\end{lemma}

\noindent
\proof
\setcounter{part_counter}{0}
Let $\mathfrak{X}= \args{\vec{X}_{\!1}, \ldots, \vec{X}_{\!n}}$ be a multiple alignment of sample $\S{S}_n = \args{X_1, \ldots, X_n} \in \S{P}^n$. We show that $\vec{M}_{\!\mathfrak{X}} \in \S{X}$.
\begin{part}
We first show that $\vec{M}_{\!\mathfrak{X}} \in [0,1]^{\ell \times m}$.
\end{part}

\begin{part}
We have
\begin{align*}
\vec{M}_{\!\mathfrak{X}}^T\vec{1}_\ell 
&= \argsS{\frac{1}{n} \sum_{i=1}^n \vec{X}_{\!i}}{^T}\vec{1}_\ell
= \frac{1}{n} \sum_{i=1}^n \vec{X}_{\!i}^T\vec{1}_\ell
= \frac{1}{n} \sum_{i=1}^n \vec{1}_m
= \vec{1}_m.
\end{align*}
This shows that $\vec{M}_{\!\mathfrak{X}} \in \S{X}$.
\end{part}
\qed

\subsection{Proof of Theorem \ref{theorem:equivalence:Fn-gn}}

We assume that $\S{S}_n = \args{X_1, \ldots, X_n} \in \S{P}^n$ is a sample of partitions.

\begin{lemma}
Consider the functions
\begin{align*}
f_n\!\args{\mathfrak{X}} &= \frac{1}{n}\sum_{i=1}^n \normS{\vec{X}_{\!i} - \vec{M}_{\!\mathfrak{X}}}{^2}\\
g_n\!\args{\mathfrak{X}} &= \frac{1}{n^2}\sum_{i=1}^n \sum_{j=1}^n \normS{\vec{X}_{\!i} - \vec{X}_{\!j}}{^2}
\end{align*}
defined on $\S{A}_n$. Then we have $\S{M}(f_n) = \S{M}(g_n)$.
\end{lemma}

\noindent
\proof
It is sufficient to show that minimizing $f_n$ and $g_n$ is equivalent to maximizing the function 
\[
h_n\!\args{\mathfrak{X}} = \sum_{i=1}^n\inner{\vec{X}_{\!i},\vec{M}_{\!\mathfrak{X}}}.
\]
By equivalence we mean that any minimizer of $f_n$ (and $g_n$) is a maximizer of $h_n$ and vice versa.  
\setcounter{part_counter}{0}
\begin{part}
 We can equivalently rewrite $f_n$ as follows:
\begin{align*}
f_n\!\args{\mathfrak{X}} &= \frac{1}{n}\sum_{i=1}^n \normS{\vec{X}_{\!i} - \vec{M}_{\!\mathfrak{X}}}{^2}\\
&= \frac{1}{n}\sum_{i=1}^n \normS{\vec{X}_{\!i}}{^2} \;-\;\; \frac{2}{n}\sum_{i=1}^n\inner{\vec{X}_{\!i},\vec{M}_{\!\mathfrak{X}}} \;+\; \normS{\vec{M}_{\!\mathfrak{X}}}{^2}\\
&= \underbrace{\frac{1}{n}\sum_{i=1}^n \normS{\vec{X}_{\!i}}{^2}}_{=\, \text{const}} \;-\;\; \frac{1}{n}\underbrace{\sum_{i=1}^n\inner{\vec{X}_{\!i},\vec{M}_{\!\mathfrak{X}}} }_{= \,h_n\!\args{\mathfrak{X}}}.
\end{align*}
 From Prop.~\ref{prop:N(X)} follows that the first sum is independent of the choice of representation and therefore constant. The last equation follows from
\begin{align*}
\normS{\vec{M}_{\!\mathfrak{X}}}{^2} 
&= \inner{\vec{M}_{\!\mathfrak{X}}, \vec{M}_{\!\mathfrak{X}}}
=  \inner{\frac{1}{n} \sum_{i=1}^n\vec{X}_{\!i}, \vec{M}_{\!\mathfrak{X}}}
= \frac{1}{n} \sum_{i=1}^n \inner{\vec{X}_{\!i}, \vec{M}_{\!\mathfrak{X}}}.
\end{align*}
This proves that minimizing $f_n$ is equivalent to maximizing $h_n$. 
\end{part}
\begin{part}
We have
\begin{align*}
g_n\!\args{\mathfrak{X}} &= \frac{1}{n^2}\sum_{i=1}^n \sum_{j=1}^n \normS{\vec{X}_{\!i} - \vec{X}_{\!j}}{^2}\\
&=\frac{1}{n^2}\sum_{i=1}^n \sum_{j=1}^n \args{\normS{\vec{X}_{\!i}}{^2} - 2\inner{\vec{X}_i, \vec{X}_{\!j}} + \normS{\vec{X}_{\!j}}{^2}}\\
&=\frac{1}{n}\sum_{i=1}^n \normS{\vec{X}_{\!i}}{^2} \;-\;\; \frac{2}{n^2}\sum_{i=1}^n \inner{\vec{X}_i, \sum_{j=1}^n  \vec{X}_{\!j}} \;+\;\;  \frac{1}{n} \sum_{j=1}^n \normS{\vec{X}_{\!j}}{^2}\\
&=\underbrace{\frac{1}{n}\sum_{i=1}^n \normS{\vec{X}_{\!i}}{^2}}_{= \,\text{const}} \;-\;\; \frac{2}{n}\sum_{i=1}^n \inner{\vec{X}_i, \vec{M}_{\mathfrak{X}}} \;+\;\;  \underbrace{\frac{1}{n} \sum_{j=1}^n \normS{\vec{X}_{\!j}}{^2}}_{= \,\text{const}}\\
\end{align*}
Again from Prop.~\ref{prop:N(X)} follows that the first and third sum are constant. This shows that minimizing $g_n$ is equivalent to maximizing $h_n$.
 \qed
\end{part}

\bigskip

\noindent
We define the set 
\[
\S{M_X}\!\args{F_n} = \cbrace{\vec{M}\in M \,:\, M \in \S{M}\!\args{F_n}}
\]
consisting of all representations that can be derived from the mean partition set $ \S{M}\!\args{F_n}$. The following relationship between the minima of $f_n$ and $F_n$ holds:
\begin{lemma}\label{lemma:fn-Fn}
For any sample $\S{S}_n \in \S{P}^n$, the map
\[
\phi:\S{M}\!\args{f_n} \rightarrow \S{M_X}\!\args{F_n}, \quad \mathfrak{X} \mapsto \vec{M}_{\!\mathfrak{X}}
\]
is surjective.
\end{lemma}

\noindent
\proof
We first show $F_n(M) = f_n\!\args{\mathfrak{X}}$ for all minimizers $\mathfrak{X} = \args{\vec{X}_1, \ldots, \vec{X}_n}$ of $f_n$ such that the mean $\vec{M}_{\!\mathfrak{X}}$ of $\mathfrak{X}$ is a representation of partition $M$. The second part shows that $M$ is a mean partition. This proves that the image of $\phi$ is contained in $\S{M_X}\!\args{F_n}$. Finally, the third part shows that $\phi$ is surjective.

\setcounter{part_counter}{0}
\begin{part}
Let $\mathfrak{X} = \args{\vec{X}_1, \ldots, \vec{X}_n}$ be a minimizer of $f_n$. Then the mean $\vec{M}_{\!\mathfrak{X}}$ of $\mathfrak{X}$ represents the partition $M = \pi\args{\vec{M}_{\!\mathfrak{X}}}$.  Observe that
\begin{align}\label{eq:proof:lemma:fn-Fn:01}
F_n(M) &= \frac{1}{n} \sum_{i=1}^n \delta\args{X_i, M}{^2} 
= \frac{1}{n} \sum_{i=1}^n \min_{\vec{X}'_i \in X_i} \normS{\vec{X}'_{\!i} - \vec{M}_{\!\mathfrak{X}}}{^2} = \hat{f}_n\args{\vec{M}_{\!\mathfrak{X}}}
\end{align}
Then by definition, we have
\begin{align}\label{eq:proof:lemma:fn-Fn:02}
\hat{f}_n\args{\vec{M}_{\!\mathfrak{X}}} \leq  \frac{1}{n} \sum_{i=1}^n\normS{\vec{X}_{\!i} - \vec{M}_{\!\mathfrak{X}}}{^2} = f_n\!\args{\mathfrak{X}}.
\end{align}
We show that $\hat{f}_n\args{\vec{M}_{\!\mathfrak{X}}} =  f_n\!\args{\mathfrak{X}}$. Let $\vec{X}'_{\!i} \in X_i$ be representations in optimal position with $\vec{M}_{\!\mathfrak{X}}$. Then we have
\begin{align*}
 \hat{f}_n\args{\vec{M}_{\!\mathfrak{X}}} &= \frac{1}{n} \sum_{i=1}^n \normS{\vec{X}'_{\!i} - \vec{M}_{\!\mathfrak{X}}}{^2}.
\end{align*}
The unique minimum of the function
\[
h_n(\vec{Z}) = \frac{1}{n} \sum_{i=1}^n \normS{\vec{X}'_{\!i} - \vec{Z}}{^2}.
\]
is the mean $\vec{M}_{\!\mathfrak{X}'}$ of the multiple alignment $\mathfrak{X}' = \args{\vec{X}'_{\!1}, \ldots, \vec{X}'_{\!n}}$. This gives
\begin{align}\label{eq:proof:lemma:fn-Fn:03}
\hat{f}_n\args{\vec{M}_{\!\mathfrak{X}}} \geq h_n\!\args{\vec{M}_{\!\mathfrak{X}'}} = f_n\!\args{\mathfrak{X}'} \geq f_n\!\args{\mathfrak{X}}.
\end{align}
The last inequality follows, because $\mathfrak{X}$ is a minimizer of $f_n$. Combing Eq.~\eqref{eq:proof:lemma:fn-Fn:02} and \eqref{eq:proof:lemma:fn-Fn:03} yields $\hat{f}_n\args{\vec{M}_{\!\mathfrak{X}}} = f_n\!\args{\mathfrak{X}}$. In summary, from Eq.~\eqref{eq:proof:lemma:fn-Fn:01}--\eqref{eq:proof:lemma:fn-Fn:03} follows
\begin{align}\label{eq:proof:lemma:fn-Fn:04}
F_n(M) = \hat{f}_n\args{\vec{M}_{\!\mathfrak{X}}} = f_n\!\args{\mathfrak{X}}.
\end{align}
\end{part}

\begin{part}
Suppose that $M$ is not a mean partition. Then for any mean partition $M'$, we have $F_n(M) > F_n(M')$. Let $\vec{M}'$ be a representation of $M'$. By Theorem \ref{theorem:MPT} there is a multiple alignment $\mathfrak{X}' = \args{\vec{X}'_1, \ldots, \vec{X}'_n}$ in optimal position with $\vec{M}'$ such that $\vec{M}' = \vec{M}_{\!\mathfrak{X}'}$. Applying Eq.~\eqref{eq:proof:lemma:fn-Fn:04} yields
\[
f_n\!\args{\mathfrak{X}} = F(M) > F_n(M') = f_n\!\args{\mathfrak{X}'},
\]
which contradicts our assumption that $\mathfrak{X}$ is a minimizer of $f_n$. Hence, $M$ is a mean partition. This shows that the image of $\phi$ is contained in $\S{M_X}\!\args{F_n}$.
\end{part}

\begin{part}
Let $\vec{M} \in \S{M}_n$ be a representation of a mean partition $M \in \S{M}\args{F_n}$. By Theorem \ref{theorem:MPT} there is a multiple alignment  $\mathfrak{X} = \args{\vec{X}_1, \ldots, \vec{X}_n}$ in optimal position with $\vec{M}$ such that $\vec{M} = \vec{M}_{\!\mathfrak{X}}$. Applying Eq.~\eqref{eq:proof:lemma:fn-Fn:04}  yields $F_n(M) = f_n\!\args{\mathfrak{X}}$. It remains to show that $\mathfrak{X}$ is a minimizer of $f_n$. Suppose that there is a multiple alignment $\mathfrak{X}'$ such that $f_n(\mathfrak{X}') < f_n(\mathfrak{X})$. Let $M'$ be the partition represented by the mean $\vec{M}_{\mathfrak{X}'}$ of $\mathfrak{X}'$. Then we have
\[
F_n(M') = f_n\!\args{\mathfrak{X}'} < f_n(\mathfrak{X}) = F_n(M),
\]
which contradicts our assumption that $M$ is a mean partition. This shows that $\phi$ is surjective. \qed
\end{part}

\subsubsection*{Proof of Theorem \ref{theorem:equivalence:Fn-gn}}

Recall that $\pi: \S{M_X}\!\args{F_n}  \rightarrow \S{M}\!\args{F_n}$ denotse the natural projection that sends matrices $\vec{M}$ to partitions $M$ they represent. By construction $\pi$ is surjective. Hence, the assertion of Theorem \ref{theorem:equivalence:Fn-gn} follows from Lemma \ref{lemma:fn-Fn}.

\end{appendix}


\begin{thebibliography}{00}
\setlength{\parskip}{0pt}
\setlength{\itemsep}{0pt plus 0.3ex}
\small
\bibitem{Bhattacharya2012}
A.~Bhattacharya and R.~Bhattacharya.
\newblock \emph{Nonpartmetric Inference on Manifolds with Applications to Shape Spaces}. 
\newblock Cambridge University Press, 2012.

\bibitem{Bredon1972}
G. E. Bredon.
\newblock{Introduction to Compact Transformation Groups}.
\newblock Elsevier, 1972.

\bibitem{Dimitriadou2002}
E. Dimitriadou, A. Weingessel, and K. Hornik.
\newblock A Combination Scheme for Fuzzy Clustering.
\newblock \emph{Advances in Soft Computing}, 2002.

\bibitem{Domeniconi2009}
C.~Domeniconi and M.~Al-Razgan.
\newblock Weighted cluster ensembles: Methods and analysis.
\newblock \emph{ACM Transactions on Knowledge Discovery from Data}, 2(4):1--40, 2009.

\bibitem{Filkov2004}
V.~Filkov and S.~Skiena. 
\newblock Integrating microarray data by consensus clustering.
\newblock \emph{International Journal on Artificial Intelligence Tools}, 13(4):863--880, 2004.

\bibitem{Franek2014}
L. Franek and X. Jiang.
\newblock Ensemble clustering by means of clustering embedding in vector spaces.
\newblock \emph{Pattern Recognition}, 47(2):833--842, 2014.

\bibitem{Frechet1948}
M.~Fr\'{e}chet.
\newblock Les \'el\'ements al\'eatoires de nature quelconque dans un espace distanci\'e.
\newblock \emph{Annales de l'institut Henri Poincar\'e}, 215--310, 1948.

\bibitem{Ghaemi2009}
R. Ghaemi, N. Sulaiman, H. Ibrahim, and N. Mustapha.
\newblock A Survey: Clustering Ensembles Techniques.
\newblock \emph{Proceedings of World Academy of Science, Engineering and Technology}, 38:644--657, 2009.

\bibitem{Gionis2007}
A.~Gionis, H.~Mannila, and P.~Tsaparts.
\newblock Clustering aggregation.
\newblock \emph{ACM Transactions on Knowledge Discovery from Data}, 1(1):341--352, 2007.

\bibitem{Gordon2001}
A.D.~Gordon and M-~Vichi. 
\newblock Fuzzy partition models for fitting a set of partitions. 
\newblock \emph{Psychometrika}, 66(2):229--247, 2001.

\bibitem{Gusfield1997}
D. Gusfield.
\newblock \emph{Algorithms on strings, trees and sequences: computer science and computational biology}. 
\newblock Cambridge University Press, 1997.

\bibitem{Hornik2008}
K.~Hornik and W.~B\"ohm. 
\newblock Hard and soft Euclidean consensus partitions. 
\newblock \emph{Data Analysis, Machine Learning and Applications}, 2008.

\bibitem{Jain2015}
B.J.~Jain.
\newblock Geometry of Graph Edit Distance Spaces.
\newblock \emph{arXiv: 1505.08071}, 2015.

\bibitem{Jain2015a}
B.J.~Jain.
\newblock Properties of the Sample Mean in Graph Spaces and the Majorize-Minimize-Mean Algorithm.
\newblock \emph{arXiv:1511.00871}, 2015.

\bibitem{Jain2015c}
B.J.~Jain.
\newblock Asymptotic Behavior of Mean Partitions in Consensus Clustering.
\newblock \emph{arXiv:1512.06061}, 2015

\bibitem{Jain2016}
B.J.~Jain.
\newblock Homogeneity of Cluster Ensembles.
\newblock \emph{arXiv:1602.02543}, 2016.

\bibitem{Li2007}
T.~Li, C.~Ding and M.I.~Jordan.
\newblock Solving consensus and semi-supervised clustering problems using nonnegative matrix factorization.
\newblock \emph{IEEE International Conference on Data Mining}, 2007.

\bibitem{Luxburg2010}
U.~von Luxburg. 
\newblock \emph{Clustering stability: An overview}. 
\newblock Now Publishers Inc., 2010.

\bibitem{Ratcliffe2006}
J.G.~Ratcliffe.
\newblock \emph{Foundations of Hyperbolic Manifolds}.
\newblock Springer, 2006.

\bibitem{Strehl2002}
 A. Strehl and J. Ghosh.
 \newblock Cluster Ensembles -- A Knowledge Reuse Framework for Combining Multiple Partitions.
 \newblock \emph{Journal of Machine Learning Research}, 3:583--617, 2002.
 
\bibitem{Topchy2005}%
A.P.~Topchy, A.K. Jain, and W.~Punch.
\newblock Clustering ensembles: Models of consensus and weak partitions.
\newblock \emph{IEEE Transactions in Pattern Analysis and Machine Intelligence}, 27(12):1866--1881, 2005.

\bibitem{VegaPons2010}
S.~Vega-Pons, J.~Correa-Morris and J.~Ruiz-Shulcloper.
\newblock Weighted partition consensus via kernels.
\newblock \emph{Pattern Recognition}, 43(8):2712--2724, 2010.

\bibitem{VegaPons2011}
S.~Vega-Pons and J.~Ruiz-Shulcloper.
\newblock A survey of clustering ensemble algorithms.
\newblock \emph{International Journal of Pattern Recognition and Artificial Intelligence}, 25(03):337--372, 2011.

\end{thebibliography}
\end{document}